\documentclass[runningheads]{llncs}

\usepackage{eccv}

\usepackage{eccvabbrv}

\usepackage{graphicx}
\usepackage{booktabs}
\usepackage{authblk}

\usepackage[accsupp]{axessibility}  

\usepackage{hyperref}

\usepackage{orcidlink}

\begin{document}

\title{Continual Learning for Remote Physiological Measurement: Minimize Forgetting and Simplify Inference} 

\titlerunning{Continual Learning for Remote Physiological Measurement}

\author{Qian Liang \and Yan Chen \and Yang Hu\thanks{Corresponding author}}

\authorrunning{Q.~Liang et al.}

\institute{University of Science and Technology of China\\ 
\email{qianliang@mail.ustc.edu.cn, \{eecyan,eeyhu\}@ustc.edu.cn}}

\maketitle

\begin{abstract}
  Remote photoplethysmography (rPPG) has gained significant attention in recent years for its ability to extract physiological signals from facial videos. While existing rPPG measurement methods have shown satisfactory performance in intra-dataset and cross-dataset scenarios, they often overlook the incremental learning scenario, where training data is presented sequentially, resulting in the issue of catastrophic forgetting. Meanwhile, most existing class incremental learning approaches are unsuitable for rPPG measurement. In this paper, we present a novel method named ADDP to tackle continual learning for rPPG measurement. We first employ adapter to efficiently finetune the model on new tasks. Then we design domain prototypes that are more applicable to rPPG signal regression than commonly used class prototypes. Based on these prototypes, we propose a feature augmentation strategy to consolidate the past knowledge and an inference simplification strategy to convert potentially forgotten tasks into familiar ones for the model. To evaluate ADDP and enable fair comparisons, we create the first continual learning protocol for rPPG measurement. Comprehensive experiments demonstrate the effectiveness of our method for rPPG continual learning. Source code is available at \url{https://github.com/MayYoY/rPPGDIL}.
  \keywords{Remote physiological measurement \and Continual learning  \and Domain prototype}
\end{abstract}

\section{Introduction}
\label{sec:intro}

The measurement of physiological signals, heart rate (HR) and heart rate variability (HRV) for instance, holds significant importance across various fields, such as medical diagnosis and emotion recognition~\cite{benezeth2018remote, yu2021facial, mcduff2023camera}. Traditional physiological measurement relies on specialized contact sensors, which are not only inconvenient to use but also potentially uncomfortable for subjects. In contrast, remote photoplethysmography (rPPG) can extract physiological signals associated with heartbeats by detecting periodic color changes in facial skin. Due to its capability of facilitating contactless physiological measurements using a common camera, rPPG has gained growing attention in recent years~\cite{poh2010non, de2013robust, lu2023neuron, du2023dual, liu2020multi, niu2020video, lu2021dual}.

Although existing deep learning-based rPPG approaches~\cite{yu2022physformer, liu2020multi, niu2020video, chen2018deepphys, yu2019remote} have achieved impressive performance, most of them only focus on the setting where training data is collected in a single session and the model is trained only once. However, developing a robust model for real-world applications necessitates a large and diverse dataset that encompasses various scenarios. Collecting such a dataset in a single session can be extremely challenging and nearly impossible. In practice, data from diverse scenarios is often collected gradually and the model is updated accordingly each time a new set of training data becomes available. This setting of training a model sequentially on a series of datasets (referred to as "tasks" interchangeably in this paper) is known as continual learning. Due to its great practical value, continual learning has attracted significant attention recently. However, this setting has not been explored by the rPPG researches.

\begin{figure}[tb]
  \centering
  \includegraphics[width=12cm]{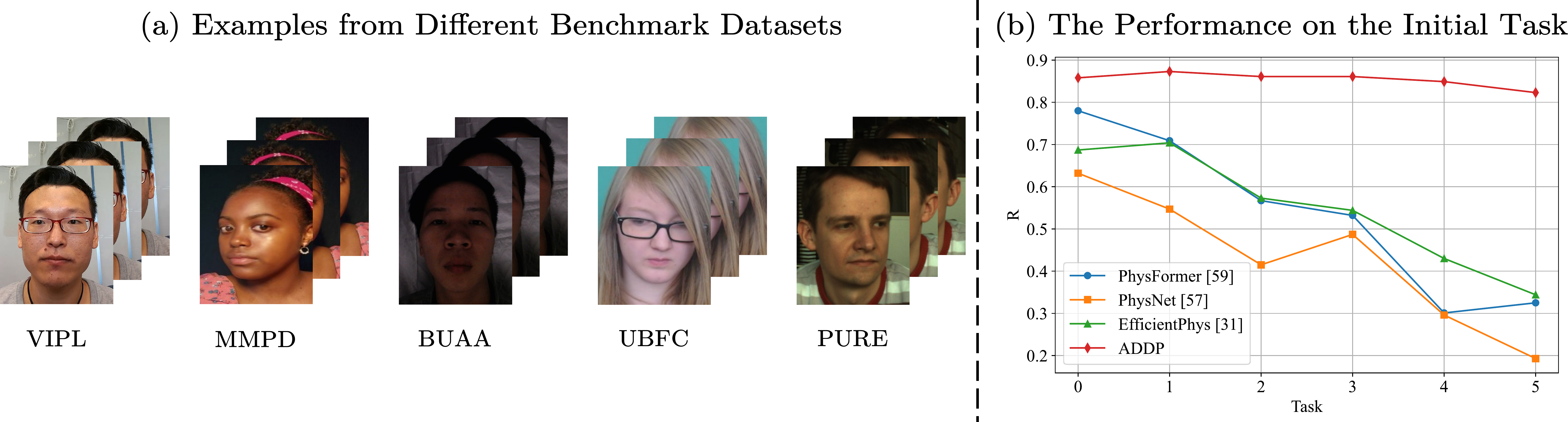}
  \caption{(a) Example sequences from popular benchmark rPPG datasets. Samples from different datasets have different backgrounds, skin colors, head motions, lighting conditions, \etc. (b) The performance of mainstream rPPG methods on the initial task during the incremental learning process. Previous methods exhibit obvious catastrophic forgetting while our method (ADDP) effectively alleviates this phenomenon.
  }
  \label{fig:figure1}
\end{figure}

In the context of rPPG measurement, factors such as lighting condition, skin color and motion can lead to distribution shifts across different datasets (See \cref{fig:figure1} (a)). Consequently, the sequential training on a series of rPPG datasets can be framed as domain incremental learning, which is the primary focus of this work. The key challenge of incremental learning is catastrophic forgetting~\cite{goodfellow2013empirical, robins1995catastrophic}, where the model tends to forget the past knowledge after finetuning on new tasks, resulting in a significant decrease in performance on previous data. To investigate whether this phenomenon occurs in rPPG measurement, we conduct a continual learning evaluation and \cref{fig:figure1} (b) shows the performance of three state-of-the-art rPPG methods on the initial task after learning new tasks sequentially. It can be observed that these methods experience significant performance degradation during the continual learning process, indicating the models' struggle to retain past knowledge and the occurrence of catastrophic forgetting.

To alleviate catastrophic forgetting, many methods preserve samples from old tasks and replay them during the training on new tasks. Although these methods demonstrate impressive performance, applying them to rPPG measurement is not practical. RPPG methods typically require long videos as input, with clip lengths exceeding 100 frames. In such cases, replaying previous samples could result in an extremely heavy storage burden. Furthermore, the stored facial videos involve subjects' privacy information, which is undesirable in real-world applications. By contrast, rehearsal-free algorithms retain no samples from previous tasks and are more practical. These methods commonly relies on class prototypes to preserve decision boundary~\cite{malepathirana2023napa, zhu2021prototype, zhu2021class}, which is infeasible for rPPG measurement as both of the ground truth heart rate and rPPG signal are continuous. Another promising approach involves leveraging the generalizability of pre-trained models and training task-specific prompts~\cite{smith2023coda, wang2022s, wang2022dualprompt}. However, our empirical results demonstrate that prompt-based methods are not directly applicable for rPPG measurement (refer to \cref{sec:overall-framework} and \cref{sec:effect-backbone} for details). To circumvent these challenges, we propose a novel rehearsal-free method based on \textbf{AD}apter and \textbf{D}omain \textbf{P}rototypes (\textbf{ADDP}) as follows:

\begin{itemize}
    \item Adapter-based Finetuning: In order to improve the model's stability and adapt it to new tasks more efficiently, we utilize adapter~\cite{houlsby2019parameter} to finetune the frozen backbone. Additionally, we design a Difference Normalization (DiffNorm) module for the backbone, which can effectively integrate dynamic and appearance features of the input videos.
    \item Prototype-based Augmentation: Although adapter finetuning minimizes the changes of parameters, continually finetuning a single group of adapters inevitably leads to forgetting. For consolidating the past knowledge, we propose to extract domain prototypes, including style prototypes and noise prototypes, of previous tasks and employ them to augment new samples during the training on new tasks.
    \item Prototype-based Inference Simplification: Drawing inspiration from the fact that people tend to simplify unfamiliar tasks by relating them with familiar ones, we propose to utilize the style prototype most familiar to the model to transfer the style of test samples, which enables the model to solve the inference problem in its most proficient manner.
\end{itemize}

The contributions of this work are 1) To the best of our knowledge, our approach is the first to explore domain incremental learning for rPPG measurement. 2) To tackle this problem, we first employ adapter to finetune the backbone. Additionally, we design domain prototypes to replace class prototypes for such regression problem and introduce prototype-based augmentation to alleviate catastrophic forgetting. Furthermore, we leverage the style prototypes to simplify the inference. 3) We establish a practical domain incremental learning benchmark for rPPG measurement. Extensive experiments show the superiority of the proposed method.

\section{Related work}

\subsection{Remote Physiological Measurement}

Remote physiological measurement relies on detecting color changes in facial skin caused by heartbeats to estimate vital signals like heart rate. Traditional methods mainly leverage blind source separation or color space transformations to extract rPPG signals that possess a high signal-to-noise ratio (SNR)~\cite{poh2010non, wang2014exploiting, de2013robust, de2014improved}. However, these methods heavily rely on prior assumptions and struggle to perform well in complex environments. Recently, deep learning approaches have been successfully used in rPPG measurement~\cite{yu2022physformer, liu2020multi, niu2020video, chen2018deepphys, yu2019remote}. They utilize 3D-CNN or modified vision transformer (ViT)~\cite{dosovitskiy2020image} to extract physiological features from raw videos~\cite{yu2022physformer, yu2019remote} or carefully designed STMaps~\cite{niu2020video, niu2019rhythmnet}. Additionally, considering that the distribution shifts between the testing and training data will limit the performance of models, some researchers have explored the problems of domain generalization and domain adaptation for rPPG measurement~\cite{lu2023neuron, du2023dual}. Unfortunately, these models still fail to generalize or adapt well to challenging domains, further prompting us to address the domain shifts in rPPG measurement through domain incremental learning.

\subsection{Continual Learing}

The primary challenge of continual learning is catastrophic forgetting. Existing methods can be broadly categorized into three classes. Architecture-based methods assign specific parameters for each task through learnable masks or expandable modules to mitigate inter-task interference~\cite{serra2018overcoming, xue2022meta, lee2020neural, rusu2016progressive}. Regularization-based methods penalize changes to important parameters or model predictions to balance the old and new tasks~\cite{aljundi2018memory, kirkpatrick2017overcoming, ahn2021ss, castro2018end, hou2018lifelong}. Rehearsal-based methods store a subset of samples from old tasks in a buffer and replay them during the training on new tasks to consolidate past knowledge~\cite{rebuffi2017icarl, hou2019learning, aljundi2019gradient, bang2021rainbow}.

Since rehearsal with stored data of old tasks will incur a large storage cost and violate data privacy, more and more researchers have started focusing on rehearsal-free continual learning. Apart from regularization and architecture-based methods, some recent approaches opt to extract feature centroids as class prototypes and utilize augmented prototypes to maintain decision boundary of previous tasks~\cite{zhu2021prototype, malepathirana2023napa}. Additionally, the emerging parameter efficient finetuning-based (PEFT) methods~\cite{wang2022learning, wang2022dualprompt, wang2022s, gao2023unified} have demonstrated the effectiveness of frozen backbones with PEFT modules for forgetting minimization. S-Prompts~\cite{wang2022s} trains task-specific prompts for each task to assign distinct subspace for different domains to tackle domain incremental learning. It further employs feature centroids to select proper prompts for inference. LAE~\cite{gao2023unified} designs a unified class incremental learning framework with PEFT modules. It calibrates the adaptation speed of PEFT modules relative to the classifiers and leverages predicted logits to ensemble an online model and an offline model. In this work, we specifically focus on domain incremental learning for rPPG measurement. Drawing inspiration from PEFT and prototype-based methods, we propose a novel and rehearsal-free method based on adapter finetuning and label-irrelevant domain prototypes that are more suitable for rPPG measurement.

\section{Preliminaries}

\begin{figure}[tb]
  \centering
  \includegraphics[width=10cm]{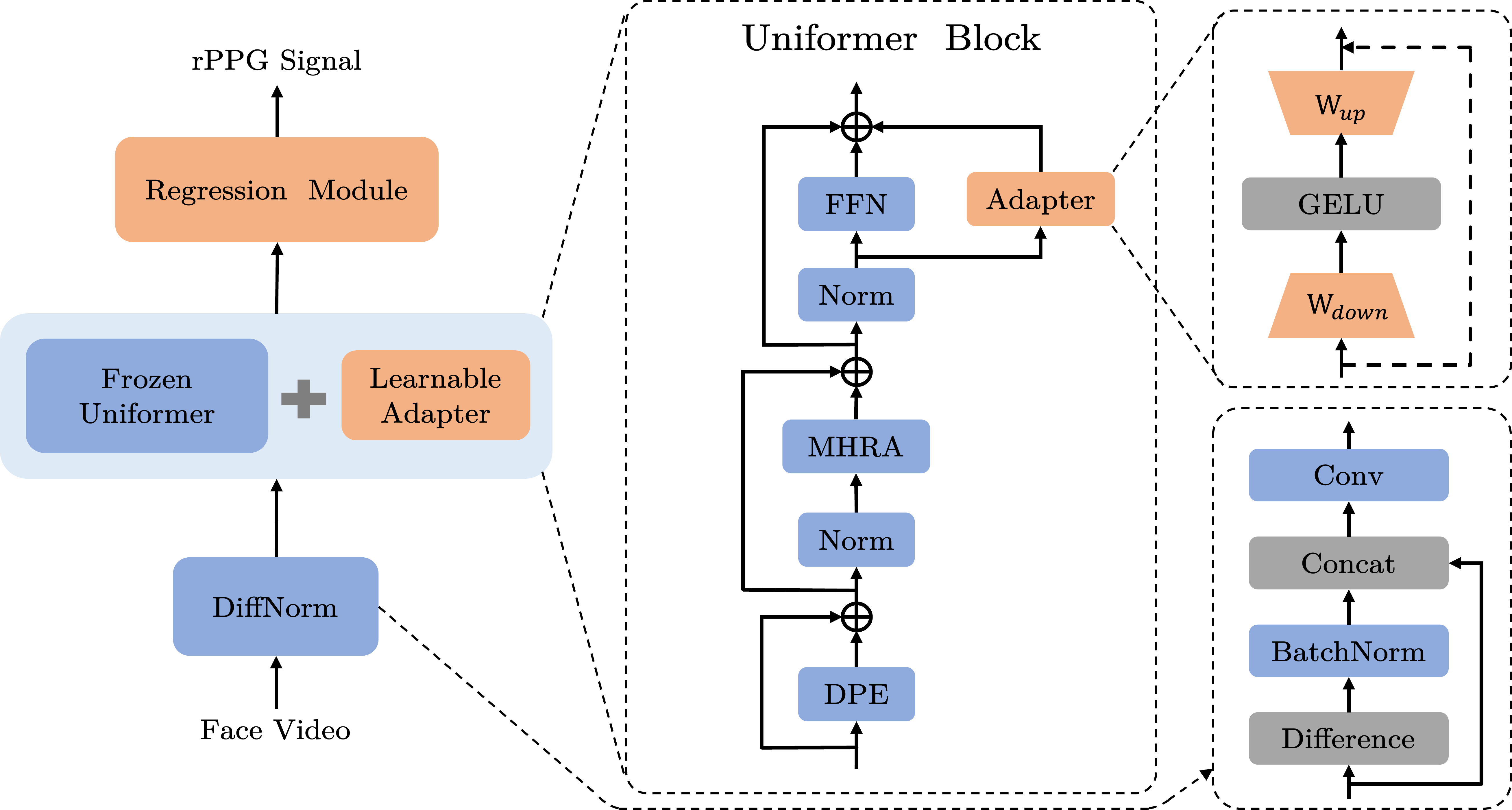}
  \caption{The architecture of our model. The base model is Uniformer which can extract the crucial local features for rPPG measurement. Our DiffNorm module can effectively fuse the appearance and dynamic features. Only the adapter and regression modules (the orange blocks) are learnable after the initial task.
  }
  \label{fig:figure2}
\end{figure}

\subsection{Problem Formulation}
In this work, we focus on domain incremental learning for rPPG measurement (rPPG DIL). Different from existing rPPG works, the model sequentially learns knowledge on different tasks with distribution shifts and is expected to perform well on all the tasks. Formally, the model $f(\cdot; \boldsymbol \theta, \boldsymbol \phi)$ is trained on tasks $\mathcal T:=\{\mathcal T_1, \mathcal T_2, \ldots, \mathcal T_n\}$ with multiple domains in sequence, where $\boldsymbol \theta, \boldsymbol \phi$ are parameters of the feature extractor and the regression module, respectively. During the inference stage, given an input facial video $\boldsymbol x\in\mathbb R^{3\times T\times H\times W}$, the model is required to accurately predict the ground truth rPPG signal $\boldsymbol y\in\mathbb R^{T}$ corresponding to $\boldsymbol x$ without knowing which task the test sample belongs to. Note that variations in illumination, subject's skin color, head motion, noises introduced by video compression and many other factors can lead to distribution shifts across rPPG datasets, and the primary challenge is to mitigate the catastrophic forgetting caused by domain shifts.

\subsection{Uniformer and Adapter}
In this part, we introduce two major parts of our feature extractor. 

\textbf{Uniformer}~\cite{li2022uniformer} is a strong spatial-temporal model for video classification. It consists of 4 stages and each stage contains a stack of Uniformer blocks. As shown in \cref{fig:figure2}, a Uniformer block mainly comprises three modules: Dynamic Position Embedding (DPE), Multi-Head Relation Aggregrator (MHRA) and Feed Forward Network (FFN). DPE leverages 3D depthwise convolution to encode position information and is friendly to arbitrary input lengths. FFN is a Multi-Layer Perceptron (MLP) same as the one used in vanilla ViT. The key module, MHRA, unifies 3D convolution and self-attention in a concise transformer format. It utilizes convolution in the first two stages to reduce spatial-temporal redundancy and capture local features while employing self-attention to model global dependency in the subsequent stages.

\textbf{Adapter}~\cite{houlsby2019parameter}, one of the most commonly used PEFT modules, can efficiently finetune a large model by inserting a small module to any layer of it. As illustrated in \cref{fig:figure2}, an adapter consists of a pair of projection matrices and a non-linear activation function (GELU in this work). The forward procedure can be formulated as follows:
\begin{equation}
    \boldsymbol x'= \boldsymbol x + \mathbf W_{up}(\operatorname{GELU}(\mathbf{W}_{down}(\boldsymbol x)))
\end{equation}
where $\boldsymbol x$ is the input of the adapter. The skip connection is optional and we do not employ it in this work.

\section{Methodology}

\subsection{Overall Framework}
\label{sec:overall-framework}
The architecture of our model is shown in \cref{fig:figure2}. Following S-Prompts~\cite{wang2022s}, we adopt the learning paradigm of freezing a pre-trained backbone and tuning a small number of learnable parameters with PEFT. In \cref{sec:effect-backbone}, we empirically validate that vanilla ViT is not suitable for our problem and that local features are crucial for rPPG measurement. Therefore, we turn to the aforementioned Uniformer~\cite{li2022uniformer}. However, the prompt finetuning~\cite{lester2021power} used by S-Prompts is not directly applicable due to the convolution layers in Uniformer.\footnote{It is feasible but less effective for rPPG measurement to forcefully utilize prefix to finetune the last two stages of Uniformer. See the supplementary material for details.} To this end, we employ the adapter finetuning and attach an adapter in parallel with the FFN for each Uniformer block. To better leverage the expanded training data and facilitate knowledge sharing among tasks, we continually finetune a single group of adapters rather than following S-Prompts to separate different PEFT modules for different tasks. Moreover, inspired by EfficientPhys~\cite{liu2023efficientphys}, we design a module called DiffNorm at the input layer, as illustrated in \cref{fig:figure2}. Building upon the normalization module of EfficientPhys, this module further utilizes convolution and concatenation to effectively integrate the appearance and dynamic features of the input videos:
\begin{equation}
    \boldsymbol x' = \operatorname{Conv}([\boldsymbol x; \operatorname{BN(\operatorname{Diff}(\boldsymbol x))}])
\end{equation}
where $x$ is the input video, $[\cdot; \cdot], \operatorname{BN}(\cdot), \operatorname{Diff}(\cdot)$ are the concatenate, batch normalization and difference operations, respectively and $\boldsymbol x'$ will be the input of the Uniformer. In summary, the feature extractor can be divided into two parts: the backbone that consists of the Diffnorm Module and the Uniformer, and the adapter for finetuning. The backbone is only trained on the initial task. Then we freeze it and only finetune the adapter and the regression module for the following tasks.

\begin{figure}[tb]
  \centering
  \includegraphics[width=11cm]{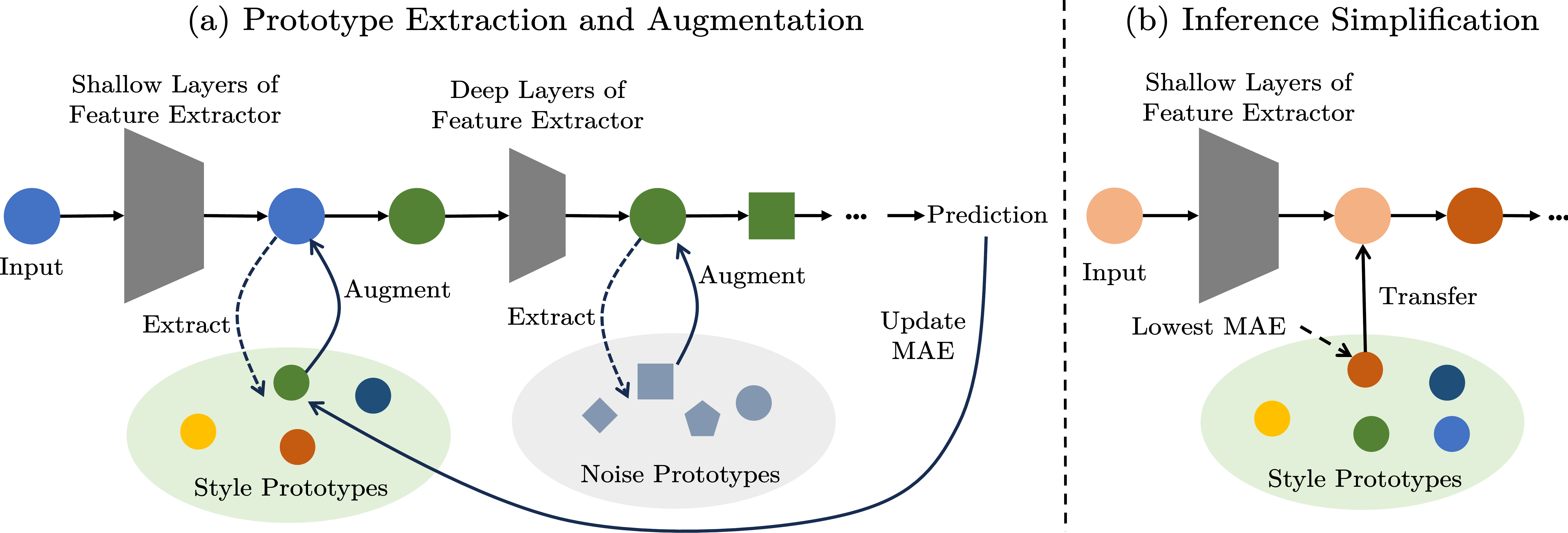}
  \caption{Overview of our domain prototype-based strategies. (a) During training stage, two types of domain prototypes, \ie style prototypes and noise prototypes, are extracted for each task. Meanwhile, we randomly select domain prototypes of previous tasks to augment the training samples and record the training MAE of the selected style prototypes. (b) In the inference stage, the style prototype with lowest MAE is selected to transfer test samples into a style that can be easily processed.
  }
  \label{fig:figure3}
\end{figure}

To address catastrophic forgetting, we propose two strategies as illustrated in \cref{fig:figure3}. Firstly, during model finetuning, we utilize prototype-based augmentation to consolidate the past knowledge. Secondly, during testing, the prototype-based inference simplification is used to convert tasks that may have been forgotten or are challenging into a form that the model is most familiar with.

\subsection{Prototype-Based Augmentation}

The primary cause of catastrophic forgetting in rPPG DIL is the variation of domain factors such as illumination condition and subject's head motion. In this work, to mitigate catastrophic forgetting, we propose to extract domain prototypes for each task and reproduce domain factors of old tasks during the training on new tasks.

Firstly, we extract the style features from training samples after training on the current task. These style features encompass crucial information of the input videos, such as illumination and the subject's skin color. Existing approaches commonly employ the channel-wise mean and variance of the feature extracted by the shallow layers of the network to represent the style distribution of the input~\cite{huang2017arbitrary, huang2023style}. In accordance with these methods, let $\boldsymbol h_{old}\in\mathbb R^{C\times T_1\times H_1\times W_1}$ denotes the shallow feature extracted by the backbone from the input $\boldsymbol x_{old}$. The style feature $\{\boldsymbol \mu_{old}, \boldsymbol \sigma_{old}\}$ corresponding to $\boldsymbol h_{old}$ can be calculated as follows:
\begin{equation}
\begin{aligned}
\boldsymbol{\mu}_{old} & =\frac{1}{T_1 H_1 W_1} \sum_{t=1}^{T_1} \sum_{h=1}^{H_1} \sum_{w=1}^{W_1} \boldsymbol{h}_{old}^{t, h, w} \\
\boldsymbol{\sigma}_{old} & =\sqrt{\frac{1}{T_1 H_1 W_1} \sum_{t=1}^{T_1} \sum_{h=1}^{H_1} \sum_{w=1}^{W_1}\left(\boldsymbol h_{old}^{t, h, w}-\boldsymbol{\mu}_{old}\right)^2}
\label{eq:style-feature}
\end{aligned}
\end{equation}

Besides rPPG features, the feature extractor also extracts noises introduced by head motion, image compression, \etc, which can not be captured by the style features. This noise may be time-varying and erroneously interpreted as rPPG features by the model. Without corresponding labels, how to disentangle them from rPPG features has always been a tough challenge in rPPG measurement. Existing methods mainly rely on adversarial learning~\cite{lu2021dual} or cross-validation~\cite{niu2020video} for noise modeling, which often introduce additional modules and incur large amount of computational and memory cost. To tackle this challenge more efficiently, we argue that high-level features extracted by a well-performing model already possess a relatively high SNR, while the aforementioned noise accounts for only a small portion of the energy in the feature maps. Therefore, taking inspiration from traditional image denoising approaches~\cite{rajwade2012image, guo2015efficient}, we can employ Singular Value Decomposition (SVD) to efficiently extract the noise features from high-level feature maps. Let $\boldsymbol z_{old}\in\mathbb R^{C_2\times T_2}$ denotes the high-level feature extracted by the backbone from input $\boldsymbol x_{old}$. The high-level semantic noise $\boldsymbol n_{old}\in\mathbb R^{C_2\times T_2}$ corresponding to $\boldsymbol z_{old}$ can be extracted as follows:
\begin{equation}
\begin{aligned}
\boldsymbol z_{old}& = \mathbf U_{old}\mathbf\Sigma_{old}\mathbf V_{old}^T\\
\mathbf M &= \operatorname{diag}(0,0,\ldots0, 1, 1,\ldots,1)\\
\boldsymbol n_{old} &= \mathbf U_{old}(\mathbf\Sigma_{old}\odot \mathbf M)\mathbf V_{old}^T
\label{eq:svd-denoise}
\end{aligned}
\end{equation}
where $\odot$ is Hadamard product and $\mathbf M\in\mathbb R^{C_2\times C_2}$ is the diagonal mask for noise extraction with first $\alpha$ diagonal elements set to zero. Finally, considering that there may be multiple domains within a single rPPG dataset as discussed in~\cite{lu2023neuron}, we apply KMeans to group the extracted style and noise features respectively, and obtain $K$ style prototypes $\{\boldsymbol \mu_k, \boldsymbol \sigma_k\}_{k=1}^K$ as well as $K$ noise prototypes $\{\boldsymbol n_k\}_{k=1}^K$ for the current task. Different from class prototypes, both of the style and noise features are highly irrelevant to the heart rate, making our domain prototypes more applicable for rPPG measurement where the labels are continuous.

With the domain prototypes of old tasks, we can replay domain factors that only occur in old tasks when training the model on new tasks. Specifically, to reproduce old styles, we employ AdaIN~\cite{huang2017arbitrary} to the shallow feature $\boldsymbol h_{new}$ of the new training sample:
\begin{equation}
\boldsymbol h_{new}^{style} = \boldsymbol\sigma_{k}\frac{\boldsymbol h_{new}- \boldsymbol \mu_{new}}{\boldsymbol\sigma_{new}} + \boldsymbol\mu_{k}
\end{equation}
where $\boldsymbol \mu_{new}$ and $\boldsymbol \sigma_{new}$ are the channel-wise mean and variance of $\boldsymbol h_{new}$ as calculated by \cref{eq:style-feature}. $\{\boldsymbol \mu_k,\boldsymbol \sigma_k\}$ is a randomly selected style prototype. For $\boldsymbol h_{new}$, this style transfer operation is applied with a probability of $p$. As for the reproduction of information in the noise prototypes, we can apply a similar operation described in \cref{eq:svd-denoise} to mix up the randomly selected noise prototype $\boldsymbol n_k$ and the training sample $\boldsymbol x_{new}$ with a probability of $p$:
\begin{equation}
\begin{aligned}
\boldsymbol z_{new} &= \mathbf U_{new}\mathbf\Sigma_{new}\mathbf V_{new}^T\\
\boldsymbol z_{new}^{noise} &= \mathbf U_{new}(\mathbf\Sigma_{new}\odot \mathbf M')\mathbf V_{new}^T + \boldsymbol n_k
\label{eq:svd-augment}
\end{aligned}
\end{equation}
where $\boldsymbol z_{new}$ is high-level feature of $\boldsymbol x_{new}$ and $\mathbf M'=\operatorname{diag}(1, 1,\ldots, 1, 0, 0,\ldots,0)$. Note that these two augmentations are performed independently and randomly, which not only serves to reinforce previous knowledge, but also ensures that the model can acquire new knowledge.

The prototype-based augmentation leverages domain prototypes extracted from the previous tasks to perform feature-level augmentation on training samples of new tasks. This strategy effectively consolidates the past knowledge by generating pseudo-old samples with domain factors of previous tasks. Meanwhile, these pseudo-old samples also serve as unseen samples for the model, enabling it to learn from a broader range of scenarios. Consequently, this strategy can also enhance the model's generalizability, which is crucial for improving performance of the model in DIL according to~\cite{cai2023rehearsal}.

\subsection{Prototype-Based Inference Simplification}

Similar to humans, models inevitably experience knowledge forgetting in the context of incremental learning. On the other hand, they also become more proficient in specific knowledge of certain tasks. Inspired by the fact that humans usually tackle difficult tasks by transforming them into familiar ones, we can utilize domain prototypes extracted during training to transform test samples into a form that the model is familiar with and can easily process. Specifically, we keep track of the training mean absolute error (MAE) associated with each style prototype used for augmenting the training samples. The style prototype with low training MAE is considered as the style that the model can easily process. During the inference stage, we leverage AdaIN to transfer the test sample $\boldsymbol x_t$ into the style that is most familiar to the model:
\begin{equation}
\boldsymbol h_{t}^{style} = \boldsymbol\sigma_{l}\frac{\boldsymbol h_{t}- \boldsymbol \mu_{t}}{\boldsymbol\sigma_{t}} + \boldsymbol\mu_{l}
\end{equation}
where $\boldsymbol h_{t}$ is the shallow feature of $\boldsymbol x_{t}$;  $\boldsymbol \mu_{t}, \boldsymbol \sigma_{t}$ are the channel-wise mean and variance of $\boldsymbol h_{t}$ and $\{\boldsymbol \mu_{l}, \boldsymbol \sigma_{l}\}$ is the style prototype with lowest training MAE.

Meanwhile, we observe that to fight against the augmentation presented in \cref{eq:svd-augment}, the model should extract more rPPG information for the feature components with large singular values. In other words, the noise augmentation can further improve the SNR of the high-level features and enhance the model's robustness to these noises. Consequently, there is no need to perform additional noise transfer during the inference stage.

\section{Experiments}

\subsection{Datasets and Evaluation Protocols}
\subsubsection{Datasets.}
\label{sec:datasets}
We use five datasets with various domain factors (lighting condition, head motion, \etc) to establish the rPPG DIL setting:

\textbf{VIPL-HR}~\cite{niu2019vipl, niu2019rhythmnet} contains 2378 RGB and 752 NIR videos recorded by 4 cameras across 9 different scenarios. In our experiment, we only use the RGB videos since the RGB modality is more commonly accessible. It is important to note that the frame rates of the videos vary, which can introduce additional distribution shifts.

\textbf{PURE}~\cite{stricker2014non} consists of 60 videos captured by an eco274CVGE camera with six different activities, namely sitting still, talking and four types of head translation and rotation.

\textbf{UBFC-rPPG}~\cite{bobbia2019unsupervised} comprises 42 uncompressed videos captured using a Logitech C920 HD Pro camera. These videos exhibit variations in lighting conditions, including varying amounts of sunlight and indoor illumination.

\textbf{BUAA-MIHR}~\cite{xi2020image} contains 165 videos recorded under varying illuminations ranging from 1.0 to 100.0 lux captured by a Logitech HD pro webcam C930E color camera. We only use data with illumination greater than or equal to 6.3 lux because underexposed videos require special algorithms that are not considered in this work.

\textbf{MMPD}~\cite{tang2023mmpd} comprises 660 mobile phone videos featuring subjects with Fitzpatrick skin types 3-6. The dataset encompasses four distinct lighting conditions, namely LED-high, LED-low, incandescent, and natural lighting. Additionally, it covers four different activities: stationary, head rotation, talking, and walking.

\subsubsection{Evaluation Protocols.}
Firstly, we would like to emphasize that compared with the other 4 folds, samples in VIPL Fold5 are found to have larger head motion and be more challenging, making it a bit more challenging for the model trained on Fold1-4 to generalize well on Fold5. Consequently, we treat Fold5 as a single task while regarding Fold1-4 and the other 4 datasets as 5 separate tasks. This results in a total of 6 tasks derived from the mentioned 5 datasets and we randomly split each task according to the subject-exclusive protocol. Meanwhile, since large-scale data is required to train a robust base model, we use VIPL Fold1-4 task as the initial task.

As for evaluation metrics, the standard deviation of the error (Std), MAE, root mean square error (RMSE) and Pearson’s correlation coefficient (R) are adopted to evaluate the performance of video-level HR estimation. The unit for Std, MAE and RMSE is beats per minute (bpm). Furthermore, to evaluate the overall performance of a model under DIL setting, we report the commonly used final incremental performance $P_N$ which can be calculated as follows:
\begin{equation}
    P_N = \frac{1}{N}\sum_{j=1}^N p_{N, j}
\end{equation}
where $p_{N,j}$ is a specific metric among Std, MAE, RMSE and R evaluated on the test set of the j-th task after learning all the N tasks. We ran all the incremental experiments 3 times with different task orders and report the mean and standard deviation of these 3 runs.

\subsection{Implementation Details}
Our proposed method is implemented using PyTorch. Following~\cite{yu2022physformer}, we first utilize the MTCNN face detector~\cite{zhang2016joint} to crop the face region in the first frame and fix the region through the following frames. Subsequently, we sample a certain video into clips with a time window of 160 with step 80 and resize them into $96\times 96$ pixels. For the ground truth rPPG signals, we employ cubic spline interpolation to align them with the corresponding videos. 

We utilize Uniformer-S~\cite{li2022uniformer} as the base model and design a regression head consist of transposed convolutional and convolutional layers. The style and noise features are extracted from the output of the second and fourth stages of Uniformer, respectively. The bottleneck ratio of adapter is set to 0.25.

To train our model, we employ the loss functions introduced in~\cite{yu2022physformer} and utilize the Adam optimizer~\cite{kingma2014adam} with an initial learning rate of 1e-4 and weight decay of 5e-5. The number of training epoch is 20 on the initial task and 10 on the following tasks. The batch size is 8 for all tasks. Hyperparameters $K, p, \alpha$ are set to 8, 0.5 and 9 respectively.\footnote{For the influence of the selection of these hyperparameters, please refer to the supplementary material for details.} Following~\cite{speth2023non, niu2019robust}, random horizontal flipping, spatially resized crop, temporal resampling and image intensity noise are used for data augmentation.

\subsection{Effectiveness of Backbone}
\label{sec:effect-backbone}
To validate the effectiveness of our backbone proposed in \cref{sec:overall-framework}, we compare our model to three baselines: vanilla ViT~\cite{dosovitskiy2020image}, PhysFormer~\cite{yu2022physformer} and vanilla Uniformer~\cite{li2022uniformer}. It is worth mentioning that we primarily focus on end-to-end transformer-based models in this work for two reasons. Firstly, non-end-to-end rPPG methods typically involve complex preprocessing procedure to generate the input STMaps, whereas end-to-end methods offer greater accessibility in real-world scenarios. Secondly, the paradigm of a frozen ViT with PEFT modules has demonstrated outstanding performance in continual learning for image classification~\cite{wang2022dualprompt, smith2023coda}.

\begin{table}[tb]
    \caption{HR estimation results of the backbone effectiveness study. We evaluate our backbone on the VIPL Fold1-4 and MMPD tasks. "EfficientNorm" denotes the normalization module in EfficientPhys~\cite{liu2023efficientphys}. The best results are in bold, and the second best results are underlined.
      }
    \label{tab:backbones}
  \centering
    \begin{tabular}{ccccccccc}
        \hline
                   & \multicolumn{4}{c}{VIPL Fold1-4} & \multicolumn{4}{c}{MMPD}  \\ \cmidrule(r){2-5} \cmidrule(r){6-9}
        Methods                     & Std$\downarrow$  & MAE$\downarrow$ & RMSE$\downarrow$ & R$\uparrow$ & Std$\downarrow$  & MAE$\downarrow$ & RMSE$\downarrow$ & R$\uparrow$ \\ \hline
        ViT~\cite{dosovitskiy2020image}        & 15.29  & 12.52  & 16.18 & 0.08 & 15.94 & 14.33 & 18.72 & 0.12 \\
        PhysFormer~\cite{yu2022physformer} & 8.09  & 6.23  & 8.62 & 0.78 & 12.14 & 8.23 & 12.16 & 0.67 \\
        Uniformer~\cite{li2022uniformer}  & 7.94  & 5.56  & 8.19 & 0.79 & \underline{9.35} & \underline{6.05} & \underline{9.45} & \underline{0.79} \\
        Uniformer + EfficientNorm  & \underline{7.85}  & \underline{5.18}  & \underline{7.85} & \underline{0.80} & 10.91 & 7.15 & 10.92 & 0.71 \\
        Uniformer + DiffNorm (ours) & \textbf{7.25} & \textbf{4.84} & \textbf{7.34} & \textbf{0.83} & \textbf{8.67} & \textbf{5.70} & \textbf{8.72} & \textbf{0.82}        \\ \hline
    \end{tabular}
\end{table}

To evaluate these models for HR estimation, we conduct intra-task evaluations on two large-scale tasks: VIPL Fold1-4 and MMPD and the results are presented in \cref{tab:backbones}. We can see that vanilla ViT completely fails to capture rPPG features for both tasks, resulting in poor performance with MAEs exceeding 10 bpm. PhysFormer, a model built upon ViT and specifically designed for rPPG measurement, introduces 3D temporal difference convolution and significantly improves the performance, achieving an MAE of 6.23 bpm on VIPL Fold1-4. The third baseline, vanilla Uniformer, further surpasses PhysFormer and get an MAE of 5.56 bpm on VIPL Fold1-4. Building upon Uniformer, we additionally design a DiffNorm module at the input layer. Despite its simplicity, this module proves to be highly beneficial, leading to a substantial reduction in Std, MAE and RMSE, as well as an improvement in R. Similar results can also be observed on MMPD. We further compare our DiffNorm module to the normalization module of EfficientPhys~\cite{liu2023efficientphys} which only extracts dynamic features from input videos. It can be observed that the normalization module in EfficientPhys is less effective on the VIPL Fold1-4 task and even detrimental to the backbone on the more challenging MMPD task due to the absence of appearance features.

To summarize, we find that vanilla self-attention mechanism is insufficient for effectively extracting rPPG features. We believe that this is because local features play a crucial role in capturing the subtle color changes caused by heartbeats. In contrast, PhysFormer and Uniformer additionally introduces 3D convolution to extract these local features. Moreover, the success of PhysFormer and our DiffNorm emphasizes the importance of difference operation and dynamic features. We use Uniformer with a DiffNorm module as the backbone in the subsequent experiments.

\subsection{Benchmark Results}
We evaluate our method (ADDP) in the rPPG DIL scenario, comparing it with 5 baselines that do not store previous samples: EWC~\cite{kirkpatrick2017overcoming}, LwF~\cite{li2017learning}, ANCL~\cite{kim2023achieving}, LAE~\cite{gao2023unified} and S-Prompts~\cite{wang2022s}. We replace the prompt pool with an adapter pool when employing S-Prompts to Uniformer. This is reasonable as prompts can be transformed into a similar form as adapters according to~\cite{he2021towards}. Similar to frequency cross entropy loss~\cite{niu2020video}, power spectral density (PSD) of the predicted rPPG signal is considered as the predicted logit for LwF and LAE. Additionally, the joint training that learns all tasks together and naive full finetuning the model without any countermeasure to forgetting are recognized as the upper and lower bounds of the performance in our rPPG DIL experiments.

\begin{table}[tb]
    \caption{HR estimation results on the rPPG DIL protocol. $\text{Std}_{N}$ represents the final incremental Std and the same applies to $\text{MAE}_{N}$, $\text{RMSE}_{N}$ as well as $\text{R}_{N}$.
      }
    \label{tab:dilhr}
    \centering
    \setlength{\tabcolsep}{3mm}{
    \begin{tabular}{cccccc}
        \hline
        Methods     & $\text{Std}_{N}\downarrow$ & $\text{MAE}_{N}\downarrow$  & $\text{RMSE}_{N}\downarrow$ & $\text{R}_{N}\uparrow$  \\ \hline
        Upper Bound & 5.25$\pm$0.09 & 3.45$\pm$0.09 & 5.31$\pm$0.24 & 0.85$\pm$0.01 \\
        Lower Bound & 8.73$\pm$0.79 & 6.54$\pm$0.27 & 9.46$\pm$0.49 & 0.69$\pm$0.02 \\ \hline
        EWC~\cite{kirkpatrick2017overcoming} & 6.88$\pm$0.89 & 4.51$\pm$0.33 & 7.02$\pm$0.86 & 0.76$\pm$0.04 \\
        ANCL-EWC~\cite{kim2023achieving} & 6.87$\pm$0.36 & 4.39$\pm$0.08 & 6.97$\pm$0.36 & 0.77$\pm$0.02 \\
        LwF~\cite{li2017learning} & 6.59$\pm$0.14 & 4.38$\pm$0.10 & 6.30$\pm$0.67 & 0.78$\pm$0.02 \\
        ANCL-LwF~\cite{kim2023achieving} & 6.37$\pm$0.28 & 4.13$\pm$0.13 & 6.12$\pm$0.25 & 0.80$\pm$0.01 \\ \hline
        LAE-online~\cite{gao2023unified} & 6.97$\pm$0.56 & 4.70$\pm$0.38 & 7.13$\pm$0.56 & 0.78$\pm$0.03 \\
        LAE-offline~\cite{gao2023unified} & 12.06$\pm$0.86 & 7.94$\pm$0.35 & 12.36$\pm$0.96 & 0.59$\pm$0.04 \\
        LAE-ensemble~\cite{gao2023unified} & 12.37$\pm$0.42 & 8.07$\pm$0.24 & 12.61$\pm$0.51 & 0.55$\pm$0.01 \\
        S-Prompts~\cite{wang2022s} & \underline{5.77}$\pm$0.18 & \underline{3.85}$\pm$0.05 & \underline{5.82}$\pm$0.19 & \underline{0.81}$\pm$0.01 \\
        \textbf{ADDP}  & \textbf{5.56}$\pm$0.04 & \textbf{3.70}$\pm$0.07 & \textbf{5.59}$\pm$0.02 & \textbf{0.83}$\pm$0.01 \\ \hline
    \end{tabular}}
\end{table}

The HR estimation results are presented in \cref{tab:dilhr}. Compared with the lower bound, both EWC and LwF effectively alleviate forgetting and achieve better last incremental performance. However, the strict stability constraints imposed on the model lead to a reduction in plasticity and forward transfer capability. ANCL further introduces an auxiliary model for plasticity regularization and slightly improves the performance. LAE is a state-of-the-art approach for class incremental learning. However, all of its three strategies (online adapter, offline adapter and experts ensemble) fail to achieve satisfactory performance in our rPPG DIL. We attribute this phenomenon to two inherent disadvantages of LAE. Firstly, the stability of LAE is highly dependent on the value of weight decay for updating the offline adapter. Secondly, the logit for ensemble prediction is hard to define in rPPG regression problem.

S-Prompts achieves the second lowest $\text{Std}_N$, $\text{MAE}_N$, $\text{RMSE}_{N}$ and the second highest $\text{R}_N$ among these baselines. Nevertheless, it still lags behind our method and the upper bound by a considerable margin. This is because S-Prompts learns each task independently, lacking knowledge transfer capability and resulting in relatively poor performance on tasks that do not correspond to the selected prompts (adapters in this work). Additionally, S-Prompts utilizes KMeans to cluster features with different labels into domain centroids, which leads to feature interference and undermines the accuracy of task prediction. By contrast, ADDP finetunes a single group of adapters to facilitate knowledge sharing and avoid task prediction, while domain prototypes further aid in consolidating past knowledge and simplifying the inference. Therefore, our method finally achieves an $\text{Std}_N$ of 5.56 bpm, an $\text{MAE}_N$ of 3.70 bpm, an $\text{RMSE}_{N}$ of 5.59 bpm and an $\text{R}_N$ of 0.83, significantly narrowing the performance gap with the upper bound.

\subsection{Ablation Study}
\label{sec:ablation-study}
The proposed ADDP consists of three main designs: style and noise prototype-based augmentation as well as inference simplification. We conduct ablation studies on these components and present the results in \cref{tab:ablate}. The first row is the baseline model that simply utilizes adapter to finetune the backbone.

\begin{table}[tb]
\caption{HR estimation results of the ablation study on rPPG DIL protocol. "Style.", "Noise." and "Sim." are style, noise prototype-based augmentation and prototype-based inference simplification respectively. "TS" means we replace our inference simplification with test-time style shifting proposed by~\cite{park2023test}.}
\label{tab:ablate}
\centering
\setlength{\tabcolsep}{2mm}{
    \begin{tabular}{ccccccccc}
        \hline
        Style. & Noise. & Sim. & $\text{Std}_{N}\downarrow$ & $\text{MAE}_{N}\downarrow$  & $\text{RMSE}_{N}\downarrow$ & $\text{R}_{N}\uparrow$    \\ \hline
        $\times$ & $\times$ & $\times$ & 7.06$\pm$0.90 & 4.59$\pm$0.53 & 7.24$\pm$0.96 & 0.78$\pm$0.03 \\
        $\surd$ & $\times$ & $\times$ & 6.68$\pm$0.67 & 4.23$\pm$0.21 & 6.83$\pm$0.60 & 0.73$\pm$0.06 \\
        $\surd$ & $\surd$ & $\times$ & 6.66$\pm$0.61 & 4.10$\pm$0.20 & 6.61$\pm$0.63 & 0.75$\pm$0.05 \\
        $\surd$ & $\surd$ & TS & \underline{5.71}$\pm$0.01 & \underline{3.91}$\pm$0.04 & \underline{5.89}$\pm$0.03 & \underline{0.81}$\pm$0.00 \\
        $\surd$ & $\surd$ & $\surd$ & \textbf{5.56}$\pm$0.04 & \textbf{3.70}$\pm$0.07 & \textbf{5.59}$\pm$0.02 & \textbf{0.83}$\pm$0.01 \\
        \hline
    \end{tabular}
}
\end{table}

We observe that the performance improves when style and noise prototype-based augmentation are employed. This demonstrates that both prototypes effectively remind the model of the domain factors of previous tasks, thereby mitigating forgetting. It is worth mentioning that we could also directly employ centroids of raw features to replay domain factors. However, this would results in feature interference among samples with different heart rates. In contrast, our style and noise features are almost irrelevant to the heart rate, allowing us to employ KMeans to group them without introducing feature interference.

We also see a significant improvement in performance when utilizing inference simplification. This makes sense because this strategy enables the transfer of potentially forgotten styles into a more familiar one for the model. We have noticed that~\cite{park2023test} also proposes a test-time style shifting (TS) strategy to address domain generalization (DG) by shifting the style of test samples to the nearest source domain. To compare our inference simplification strategy with TS, we replaced it with TS in our experiments. The results in the fourth row of \cref{tab:ablate} show that TS brings less improvement compared to our strategy. We attribute it to the difference between DG and DIL. In the case of DIL, the model has already seen almost all styles of test samples, rendering the nearest style shifting less effective during inference. Furthermore, the model may have already forgotten the nearest style, resulting in a negative impact on some tasks when employing style shifting. In contrast, we utilize the style prototype with the lowest training MAE to ensure that the model can effectively process the transferred samples.

\section{Conclusion}
This paper focuses on the domain incremental learning for rPPG measurement (rPPG DIL), which has never been explored before. To tackle this challenge, we present a practical rehearsal-free method named ADDP. Adapter finetuning is employed for efficiently adapting the model to new tasks while keeping the stability of the model. Besides, we design a simple yet effective difference normalization (DiffNorm) module to integrates the appearance and dynamic features for the backbone. To mitigate forgetting, we design rPPG-friendly domain prototypes and propose prototype-based augmentation, which generates pseudo-old samples with domain factors of previous tasks. Furthermore, we employ an inference simplification strategy to transform challenging tasks into more manageable ones. To compare our method and existing continual learning algorithms, we establish the first rPPG DIL protocol. Extensive experiments demonstrate that ADDP achieves 
satisfactory performance superior to other baselines.

\textbf{Limitations: }Firstly, ADDP relies on a well-performing backbone, which may be unavailable when the initial task is not large and diverse enough. In such cases, more tuning on the backbone may be necessary during subsequent learning. Secondly, our method has not been evaluated in the online continual learning scenario, a more realistic setting where the model learns from a single-pass data stream and can only access each batch of data once. Due to the non-stationary nature of the stream, this setting is also more challenging and may require additional designs.

\section*{Acknowledgements}
This work was supported by the National Natural Science Foundation of China (62172381).

%
%
\bibliographystyle{splncs04}

\end{document}